\pdfoutput=1
\documentclass[11pt]{article}

\usepackage[final]{acl}

\usepackage{times}
\usepackage{latexsym}
\usepackage{booktabs}
\usepackage[T1]{fontenc}
\usepackage[utf8]{inputenc}
\usepackage{microtype}
\usepackage{inconsolata}
\usepackage{graphicx}
\usepackage{geometry}
\usepackage{array}
\usepackage{caption}
\usepackage{makecell}
\usepackage{amsmath}
\usepackage{hyperref}
\usepackage{amssymb} 
\usepackage{tabularx}
\usepackage{colortbl}
\usepackage{pifont}
\usepackage{tcolorbox}
\usepackage{multirow}
\usepackage{listings}  
\usepackage{xcolor}    
\usepackage{soul}

\newcommand{\dataset}{\textsf{CollabStory}}

\title{\dataset: \\  Multi-LLM Collaborative Story Generation and Authorship Analysis}

\author{Saranya Venkatraman, Nafis Irtiza Tripto, Dongwon Lee\\    
The Pennsylvania State University, USA\\ 
\texttt{\{saranyav, nit5154, dongwon\}@psu.edu}}

\begin{document}
\maketitle
\begin{abstract}
The rise of unifying frameworks that enable seamless interoperability of Large Language Models (LLMs) has made LLM-LLM collaboration for open-ended tasks a possibility. Despite this, there have not been efforts to explore such collaborative writing. We take the next step beyond human-LLM collaboration to explore this multi-LLM scenario by generating the first exclusively LLM-generated collaborative stories dataset called {\dataset}. We focus on single-author ($N=1$) to multi-author (up to $N=5$) scenarios, where multiple LLMs co-author stories. We generate over 32k stories using open-source instruction-tuned LLMs. Further, we take inspiration from the PAN tasks \cite{pan_task} that have set the standard for human-human multi-author writing tasks and analysis. We extend their authorship-related tasks for multi-LLM settings and present baselines for LLM-LLM collaboration. We find that current baselines are not able to handle this emerging scenario. Thus, {\dataset}  is a resource that could help propel an understanding as well as the development of techniques to discern the use of multiple LLMs. This is crucial to study in the context of writing tasks since LLM-LLM collaboration could potentially overwhelm ongoing challenges related to plagiarism detection, credit assignment, maintaining academic integrity in educational settings, and addressing copyright infringement concerns. We make our dataset and code available at \texttt{\url{https://github.com/saranya-venkatraman/CollabStory}}.
\end{abstract}

\section{Introduction}
\begin{figure}[t!]
\centering
    \includegraphics[width=\columnwidth]{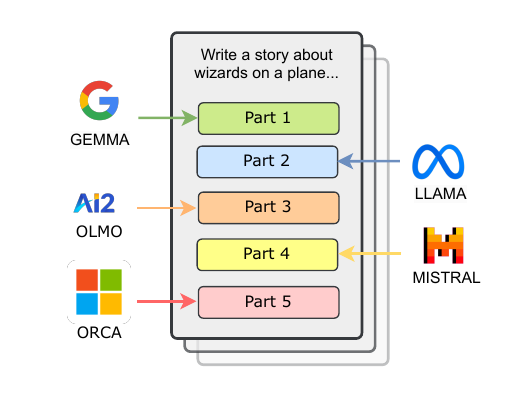}
  \caption{{\dataset} contains over $32k$ creative stories written collaboratively by up to 5 LLMs. Each story segment is generated by a single author, that then passes the narrative baton to the next, completing the storyline part by part in a sequential manner.}
  \label{fig:teaser}
\end{figure} 

\begin{table*}[ht!]
\centering
\resizebox{\textwidth}{!}{\begin{tabular}{lccccc}
\toprule
\textbf{Dataset} & \textbf{\# Stories} & \textbf{\# Authors} & \textbf{Avg \# Words} & \textbf{M-M Collaboration} & \textbf{Available} \\
\midrule
STORIUM \cite{storium} & 5,743 & 30,119 & \textasciitilde{}19k tokens & H-H \textcolor{red}{\ding{55}} & \textcolor{teal}{\ding{51}} \\
CoAuthor \cite{coauthor} & 830 & 58 & 418  & H-M \textcolor{red}{\ding{55}} & \textcolor{teal}{\ding{51}} \\
StoryWars \cite{storywars} & 40,135 & 9,494 & 367  & H-H \textcolor{red}{\ding{55}} & \textcolor{red}{\ding{55}} \\
\midrule
\rowcolor{yellow!15} \textbf{CollabStory} [Ours] & \textbf{32,503} & \textbf{5} & \textbf{725} & \textbf{M-M \textcolor{teal}{\ding{51}}} & \textcolor{teal}{\ding{51}} \\
\bottomrule
\end{tabular}}
\caption{Comparison of CollabStory with other existing collaborative creative story datasets. Here, ``M-M'' $\to$``Machine-Machine'' ,``H-H'' $\to$``Human-Human'', where ``H'' $\to$ ``Human'' and ``M'' $\to$ Machine. Ours is the largest dataset to present multi-LLM or machine-machine collaborative generation.}
\label{tab:dataset_comparison}
\end{table*}

Generative Large Language Models (LLMs) are being used more widely and becoming ubiquitous in real-world scenarios. There is particular interest in understanding the use of such LLMs in various writing tasks as writing assistants or collaborators in machine-in-the-loop settings \cite{ghostwriter, metaphorian, writingmodes, singh2023hide, yang2022ai, wordcraft2022, coauthor, clark2018creative}. So far though, this has only been explored in the case where a human is present. However given the rise of unifying frameworks that bring together and make LLMs from different sources interoperable, such as vLLM\footnote{\url{https://docs.vllm.ai/en/stable/}}, LangChain\footnote{\url{https://www.langchain.com/langchain}}, and HuggingFace\footnote{\url{https://huggingface.co/}}, the prospect of LLMs seamlessly collaborating and even handing off tasks to one another without external routing algorithms is on the horizon. This is particularly immediately possible with open-source models that are already being used by over 100K users per month (according to the number of downloads reported by HuggingFace). Despite the ease of interoperability of such LLMs, so far, automated writing assistants have been used only in collaboration with human authors or with a single LLM. Therefore, this study explores collaborative creative story-writing scenarios involving multiple LLMs, i.e. LLM-LLM collaboration.

Collaborative creative story writing entails multiple authors contributing separate segments to form a coherent storyline (see Figure \ref{fig:teaser} for our dataset schema). Although individual LLMs excel at generating story plots, collaborative writing presents unique hurdles. Models must seamlessly continue the existing storylines generated so far by other models, even if they do not align perfectly with their own language distribution. The rise of multi-agent Artificial Intelligence (AI) underscores the potential for combining the expertise of agents specialized in various tasks. While previous mixture-of-experts scenarios focused on agents proficient in task-oriented settings \cite{colab12024, colab22024agentcoord, colab32023dynamic, colab4li2023, colab52023-theory}, the emergence of LLMs conversing for continuous generative tasks in open domains is noteworthy. Imagine the possibilities when multiple LLMs collaborate; one LLM can generate compelling stories, but what if we put them together?

In this study, we attempt to address this question through a collaborative creative story-writing scenario involving multiple open-source LLMs. This is a crucial setting to study in the context of writing tasks since LLM-LLM collaboration could potentially overwhelm ongoing challenges related to plagiarism detection, credit assignment, maintaining academic integrity in educational settings, and addressing copyright infringement concerns.

We focus on single-author ($N=1$) to multi-author (up to $N=5$) scenarios, where multiple LLMs co-author creative stories. This exploration is novel, as previous studies have primarily focused on human-LLM collaboration. Towards this goal, \textbf{we generate the first multi-LLM collaborative story dataset called {\dataset} using open-source LLMs}. We select 5 frequently used LLMs (with number of downloads on HuggingFace for May 2024 provided in parenthesis): Meta's \textbf{Llama} ($>540k$ downloads, \citet{llama}), Mistral.ai's \textbf{Mistral} ($>1000k$ downloads, \citet{mistral}), Google's \textbf{Gemma} ($>180k$ downloads, \citet{gemma}), AllenAI's \textbf{Olmo} ($>26k$ downloads, \citet{olmo}) and Microsoft's \textbf{Orca} ($>22k$ downloads, \citet{orca}) to replicate a scenario in which commonly used LLMs from different organizations are being used in conjunction towards a single task. 

We demonstrate how one such dataset can be developed and the considerations involved in building an iterative Multi-LLM story-writer. We present an evaluation of the generated texts and their quality as the number of authors increases from one to five. Since this is the first such multi-LLM dataset, our evaluation provides insights into the current state of LLMs and their ability to collaborate on open-ended generation tasks. We then take inspiration from the PAN tasks \cite{pan_task} that have set the standard for multi-author writing tasks and analysis for human-human collaboration for over 15 years. We adapt their authorship-related task suite to the machine-machine settings for problems such as authorship verification and attribution, and demonstrate that current baselines are challenged by this emerging scenario. {\dataset}  is the first resource that could help propel an understanding as well as the development of new techniques to discern the use of multiple LLMs in text.  

Our work is motivated by the implications of Multi-LLM settings for different stakeholders (LLM developers, end-users) and considerations (such as credit assignment, and legality of usage) arising in the generative AI landscape. As an example, a malicious actor might assemble texts from different LLMs together in one document to evade attribution and successfully spread misinformation. Our findings elaborate on the tasks our dataset enables and emphasize the importance of tackling the incoming challenges of machine-machine collaboration.

\section{Related Work}
\paragraph{LLMs as Collaborative Writers}
LLMs are being increasingly used as writing assistants or to paraphrase, edit or enhance human-written written texts in machine-in-the-loop settings \cite{metaphorian, singh2023hide, yang2022ai, clark2018creative}. GhostWriter \cite{ghostwriter} and Wordcraft \cite{wordcraft2022} are tools that enable users to co-write stories using instructions \cite{wordcraft2022}. \citet{writingmodes} use ``writing modes'' as a control signal to better align the machine during co-writing with humans. CoAuthor positions GPT3.5 as a writing collaborator for over 50 human participants to co-write creative and argumentative stories \cite{coauthor}. 

It is interesting to consider what it means for an entity to be considered an author. LLMs do not have an innate authorship style in the way that most human writers might. It is true that an LLM likely has an emergent style that is some combination or influenced by all the different authors present in its training set. However, for the scope of this work, we turn to current literature that broadly treats authorship as a text source, hence, all the text originating from the same LLM is being treated as one group or class \cite{style1, style2, style3, style4, style5}. For example, even in the case of human writers, the same author might shift their style based on the situational context and switch between internal personas (say, a formal tone to a comical tone). Even with such style shifts, all the text that originates from the same person would still be considered to be authored by the same entity. In a similar manner, in the absence of a prompt that forces the LLM to adhere to some strict style (such as ``write a story as an 18th Century poet” or ``write a story as an angry politician”), the text being generated by LLMs seem to still have some emergent and consistent patterns that enable its detection and attribution \cite{style6, style7, style8, tripto2023ship}. Thus, in this work, when the LLM generates text from its learned distribution in the absence of any explicit prompting that might alter style in specific ways, the text being generated is considered as being a part of the LLM’s authorship. 


\paragraph{Datasets}
Despite such emerging tools, only a handful have developed datasets that can be leveraged to understand collaborative story writing. One such resource is the STORIUM dataset \cite{storium} that contains over 5k creative stories written and obtained from human-human collaboration. In the case of human-machine co-writing, CoAuthor \cite{coauthor} and CoPoet \cite{copoet} remain one of the few publically available datasets of human-machine collaborative creative story and poem writing, respectively. Beyond creative writing, \cite{hybridessays} developed the first machine-human academic essay dataset as a means to study boundary detection for academic settings. A comparison of {\dataset} with existing datasets is provided in Table \ref{tab:dataset_comparison}.

\begin{table*}[t!]
\centering
\resizebox{\textwidth}{!}{%
\begin{tabular}{ccccc|cr}
\toprule
\# Number & \# Words per Author / & \# Author Order & \# Prompts per & \# Stories & Authors & \multicolumn{1}{c}{HuggingFace distribution of LLMs used} \\
of Authors ($N$) & \# Total Words & Permutations & Author Order & & \\ \midrule
1 & 900 / 900 & 4 & 1800 & 7200 & Gemma & \texttt{google/gemma-1.1-7b-it}\footnotemark[8] \\
2 & 450 / 900 & 12 & 600 & 7200 & Llama & \texttt{meta-llama/Llama-2-13b-chat-hf}\footnotemark[9] \\
3 & 300 / 900 & 15 & 480 & 7200 & Mistral & \texttt{mistralai/Mistral-7B-Instruct-v0.2}\footnotemark[10] \\
4 & 225 / 900 & 15 & 480 & 7200 & Orca & \texttt{microsoft/Orca-2-13b}\footnotemark[11] \\
5 & 180 / 900 & 15 & 480 & 7200 & Olmo & \texttt{allenai/OLMo-7B-Instruct}\footnotemark[12] \\
\bottomrule
\end{tabular}%
}
\caption{{\dataset} creation: Summary of generation statistics}
\label{tab:story_data_collection}
\end{table*}
\footnotetext[8]{\url{https://huggingface.co/google/gemma-1.1-7b-it}}
\footnotetext[9]{\url{https://huggingface.co/meta-llama/Llama-2-13b-chat-hf}}
\footnotetext[10]{\url{https://huggingface.co/mistralai/Mistral-7B-Instruct-v0.2}}
\footnotetext[11]{\url{https://huggingface.co/microsoft/Orca-2-13b}}
\footnotetext[12]{\url{https://huggingface.co/allenai/OLMo-7B-Instruct}}

\section{Methodology}
\subsection{\dataset: Dataset Creation}
We generate a dataset of creative stories using five open-source instruction-tuned LLMs: Llama2 \cite{llama}, Olmo \cite{olmo}, Gemma \cite{gemma}, Mistral \cite{mistral} and Orca \cite{orca} (model details are provided in Table \ref{tab:story_data_collection}). The main focus of our data generation is to simulate a scenario where LLMs from different sources (organizations) collaboratively work on a storyline, handing off control of the story from one LLM to the next. The stories in our dataset vary in the number of authors/LLMs involved, from being written entirely by a single LLM to written collaboratively by between 2 to up to all 5 LLMs. In this document, we refer to each of the LLMs as ``authors". For cases where we refer to the human author, we specifically mention ``human'' author/writer. We generate our dataset by prompting various LLMs using creative writing prompts from an existing dataset called the \textbf{Writing Prompts (WP)} Dataset. The Writing Prompts Dataset was collected by \citet{writingprompts_dataset} using Reddit's \texttt{r/WritingPrompts/} forum that contains premises or prompts for stories. The WP dataset consists of a cleaned subset of story prompts and corresponding human-written stories using filtration criteria such as removal of stories that are bot-generated, less than 30 words long, contain profanity, general announcements, and so on. 
We used the test split\footnote{\url{https://www.kaggle.com/code/ratthachat/writingprompts-combine-one-line-data-for-gpt2/input?select=writingPrompts}} of this dataset as the source of prompts for LLM generated stories.

We also filter out prompts that do not have at least one corresponding human-written story that is at least 800 words long. We do this to ensure that the prompt itself does not preclude longer storylines. We chose 800 words as a criteria as a means to include stories that are slightly longer than the average of the dataset. The average length (number of words) of articles in the test set is 675.75 words. Out of 15138 total prompt-story pairs, this left us with 4623 data points. For each prompt, we divide the total goal article length (800-900 words) by the \textbf{number of authors ($N$)} to calculate the length of each part or story chunk to be written by each author, such that the writing load is distributed roughly uniformly amongst the LLM authors. We also generate different permutations of LLM authorship order such that every author can contribute to random parts of the story and we ensure that our dataset does not have any spurious correlations between LLM/author and story sections such as the beginning, or ending. For each value of the numbers of authors i.e. \( N \in \{1, 2, 3, 4, 5\} \), we generate all possible permutations of author orders. For example, for \ N = 3, two examples of author order permutations could be:
\begin{align*}
Olmo \rightarrow Mistral \rightarrow Llama \\
Gemma \rightarrow Llama \rightarrow Mistral
\end{align*}
From all such possible permutations, we sample the minimum of either total possible orders or 15 as the number of author orders. For each author order, we then generate stories using each of the prompts from a unique set of prompts per {N}. Our goal number of stories for each {N} was set to 7200 stories. A summary of the words written by each author, author order permutations, and prompts per author, as well as the pool of 5 authors and their corresponding model checkpoints used for generating all story parts is shown in Table \ref{tab:story_data_collection}. 

\subsection{LLM prompting}
For each value of \(N\), we used different prompts to generate story parts sequentially, as detailed in Table \ref{tab:prompts}. Utilizing the vLLM library\footnote{\url{https://docs.vllm.ai/en/stable/}}, we accessed and generated text from various LLMs. Initially, we conducted a pilot study to refine our prompts by generating and reviewing 100 articles. For the "Beginning" prompt, the first LLM used only the original \texttt{r/WritingPrompts/} input. For subsequent parts, we found that longer input prompts reduced story length, so we used the \textbf{Falcon.ai summarizer}\footnote{\url{https://huggingface.co/Falconsai}} to condense the story so far into under 80 words, allowing LLMs to generate longer sequences. We also included the last sentence of the story so far for smooth continuity. Prompts for different sections only varied in their instructions to begin, continue, or conclude the story. Additionally, we added an instruction to prevent LLMs from generating extraneous instructions. More details are provided in Section \ref{sec:llm_prompting}.

\begin{table}[h!]
\centering
\resizebox{0.5\textwidth}{!}{%
\begin{tabular}{m{11cm}}
\toprule
\textbf{Prompt Templates} \\
\midrule
\textit{Beginning Prompt} \\
\texttt{You are a creative story writer. Write a story that starts with the prompt \textcolor{cyan}{\{starting prompt\}} in around \textcolor{orange}{\{n\}} words. Do not add any instructions. Start the story as follows:} \\
\midrule
\addlinespace
\textit{Middle Prompt} \\
\texttt{Write \textcolor{orange}{\{n\}} words to continue this storyline: \textcolor{teal}{\{summary of story so far\}}. Continue from this sentence: \textcolor{magenta}{\{last sentence from previous part\}}} \\
\midrule
\addlinespace
\textit{Ending Prompt} \\
\texttt{Write \textcolor{orange}{\{n\}} words to conclude this storyline: \textcolor{teal}{\{summary of story so far\}}. Do not add any instructions. Continue from this sentence: \textcolor{magenta}{\{last sentence from previous part\}}} \\
\bottomrule
\end{tabular}}
\caption{Prompt templates for different parts of the story. \texttt{\textcolor{orange}{\{n\}}} here denotes the number of target words for each author.}
\label{tab:prompts}
\end{table}

\newcommand{\hly}[2][yellow!40]{ {\sethlcolor{#1} \hl{#2}} }

\begin{table*}[t!]
\centering
\resizebox{\textwidth}{!}{\begin{tabular}{|p{5cm}|p{5cm}|p{5cm}|p{5cm}|p{5cm}|}
    \toprule
        \multicolumn{1}{|c|}{\textbf{Part 1}} & \multicolumn{1}{c}{\textbf{Part 2}} & \multicolumn{1}{|c|}{\textbf{Part 3}} & \multicolumn{1}{c}{\textbf{Part 4}} & \multicolumn{1}{|c|}{\textbf{Part 5}} \\
        \midrule
            \multicolumn{1}{|c|}{\textbf{Llama}} & \multicolumn{1}{c}{\textbf{Olmo}} & \multicolumn{1}{|c|}{\textbf{Orca}} & \multicolumn{1}{c}{\textbf{Gemma}} & \multicolumn{1}{|c|}{\textbf{Mistral}} \\
        \midrule
        Time and time again, you failed. Today, you succeeded. The \hly{golden trophy} gleamed in your hands, casting a warm glow over the crowded arena. The \hly{cheers of the crowd} washed over you, a cacophony of sound that threatened to drown out your own thoughts. You couldn't believe it. You had finally done it. You had won. But the victory was tinged with a sense of sadness. You had sacrificed so much to get here. The countless hours of training, the endless miles of running, the early morning wake-up calls and the sleepless nights. It had all been worth it, you told yourself, but a part of you \hly{couldn't help but wonder if it was all worth it}. As you held the trophy aloft, the crowd cheered even louder. You closed your eyes and let the moment wash over you. This was what it was all about. This was why you had worked so hard. This was why you had pushed yourself to the limit. This was why you had failed so many times before. But deep down, you knew that this victory was not just about you. &  There was something more to the \hly{silver-haired coach}, something that you had seen the moment he had entered the arena. A quiet determination that had inspired you to push yourself beyond your limits. You had never believed in destiny or fate, but looking at the coach, you couldn't help but think that maybe there was something more to the connection between the two of you. You could feel the eyes of the crowd on you, the weight of their expectations. But you were not one to shy away from a challenge. You turned to face the coach, a determined smile on your face. "Thank you for believing in me," you said, your voice quiet but steady. The coach's face softened, and you could see the relief and pride on his face. "I never doubted your abilities, Zephyrus. You have always had the potential to be great." The words felt like a weight had been lifted off your shoulders. You knew that the road ahead would not be easy, but with the support of the coach, you felt more confident than ever. You took a step closer to the coach, reaching out to touch his hand. "I'll do everything I can to live up to your expectations," you said, your voice trembling with emotion. & \hly{The crowd roared in response}, their cheers drowning out any lingering doubts you may have had. You looked out at their faces, a sea of adoration and support, and felt a surge of determination wash over you. You would not let them down, not now, when you were so close to achieving your ultimate goal. As you took your place on the podium, you couldn't help but think back to the countless hours you had spent training, the blood, sweat, and tears you had poured into this journey. And yet, here you were, a champion, a beacon of hope for all those who had believed in you. You glanced over at your family, your friends, and your coach, their faces beaming with pride. You knew that they had sacrificed just as much as you, if not more, and you vowed to honor their faith in you by pushing yourself to new heights. With a deep breath, you stepped up to the microphone, ready to deliver your acceptance speech. You spoke from the heart, sharing your story, your struggles, and your dreams. & The words tumbled out effortlessly, revealing a vulnerability that resonated with the audience. Some wept, others roared in applause, but you felt a sense of peace that transcended all the accolades and recognition. \hly{The golden trophy felt like a symbol of your journey, not your destination.} It was the culmination of countless sleepless nights, tireless training sessions, and the unwavering support of those who believed in you. &  But the real prize, you realized, was not the gold, but the love and admiration of the people who had watched you grow into the champion you had become. You lifted the trophy high, your heart swelling with gratitude and pride. \hly{This moment, this feeling, was what made it all worth it.} As the crowd roared their approval, you took a moment to look around, taking in the faces of your family, friends, and mentors. Their eyes were filled with joy and pride, and it was in that instant that you knew you had not only won a competition, but also their hearts. With a triumphant grin, you raised the trophy once more, basking in the glory of your victory, but more importantly, the love and admiration that came with it. And in that moment, you knew that no matter what challenges came your way, you would face them with the same determination and perseverance that had brought you to this point. \\
\bottomrule
\end{tabular}}
\caption{Example of a 5-part LLM story from {\dataset} dataset for the prompt: ``\textit{Time and time again, you failed. Today, you succeeded.}'' As the story progresses, new characters are introduced (such as the coach highlighted in Part 2), with future parts linking to concepts from previous parts. In this example, the first part asks a question about the worth of the trophy, and the last part offers a completion to that sub-plot (highlighted in Part 1 and Part 5).}
\label{tab:example_stories}
\end{table*}

\subsection{Post-processing and filtering}
Though we used instruction-tuned LLMs, they do not follow instructions perfectly. Though our goal number of words per story was 800-900 words, we used the upper limit to calculate the number of words each LLM should generate. From our pilot study, we found that most LLMs were undershooting their target number of words in the instruction. We also filtered out all stories in which at least one part was under 50 words long. We also removed all extra spaces from the stories and any repetitions of the instructions in rare cases. We also filtered for some additional types of noises detailed in Section \ref{sec:data_cleaning}. After this filtration, we were left with the following number of stories per $N \in [1,5]: {7164, 7070, 6093, 6955, 5221}$ for a total of $32,503$ stories. An example of one such story from our dataset can be read in detail in Table \ref{tab:example_stories}.

\begin{figure*}[t!]
\centering
    \includegraphics[width=\textwidth]{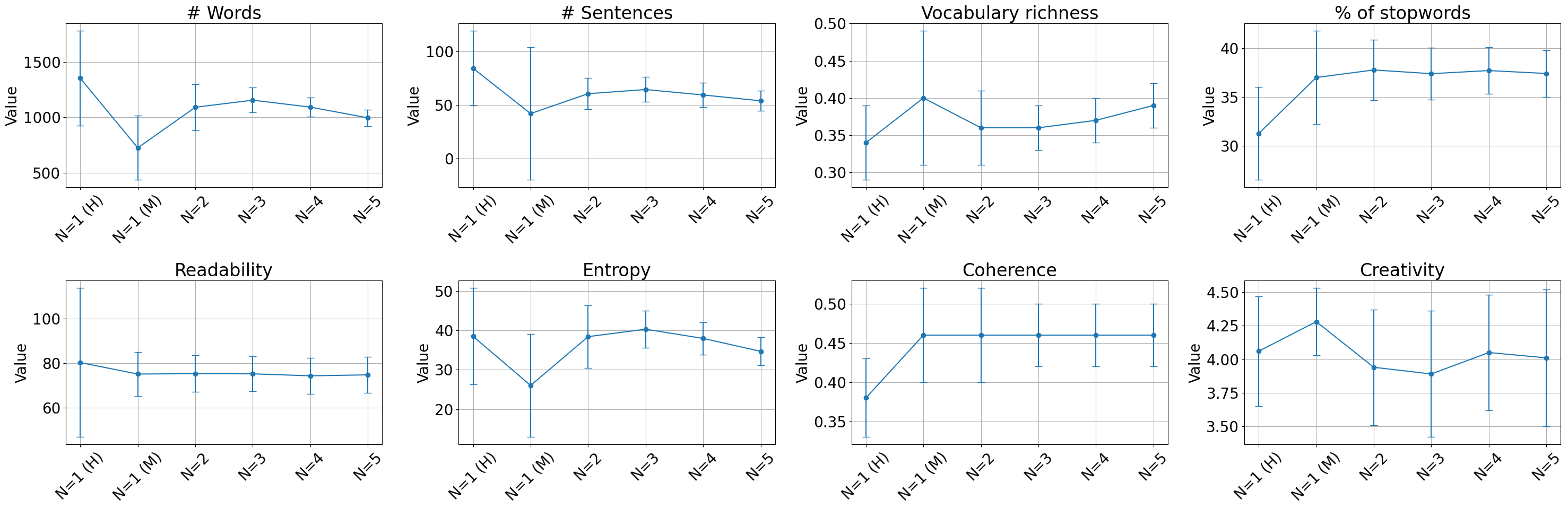}
  \caption{N on the X-axis denotes the number of authors, and N=1(H) and N=1(M) correspond to the human-written and machine-generated single-authored texts, respectively. All other texts (N >=2) are multi-LLM generated. Y-axis shows the values of the measure shown in each suplot as mentioned in the headings. For all measures, we show average and standard deviation for N going from 1 to 5. For all measures except vocabulary richness (3rd column, 1st row), increasing the number of authors (N) does not lead to statistically significant deviations from the human text distribution.}
  \label{fig:data_stats}
\end{figure*} 

\begin{figure*}[t!]
\centering
    \includegraphics[width=\textwidth]{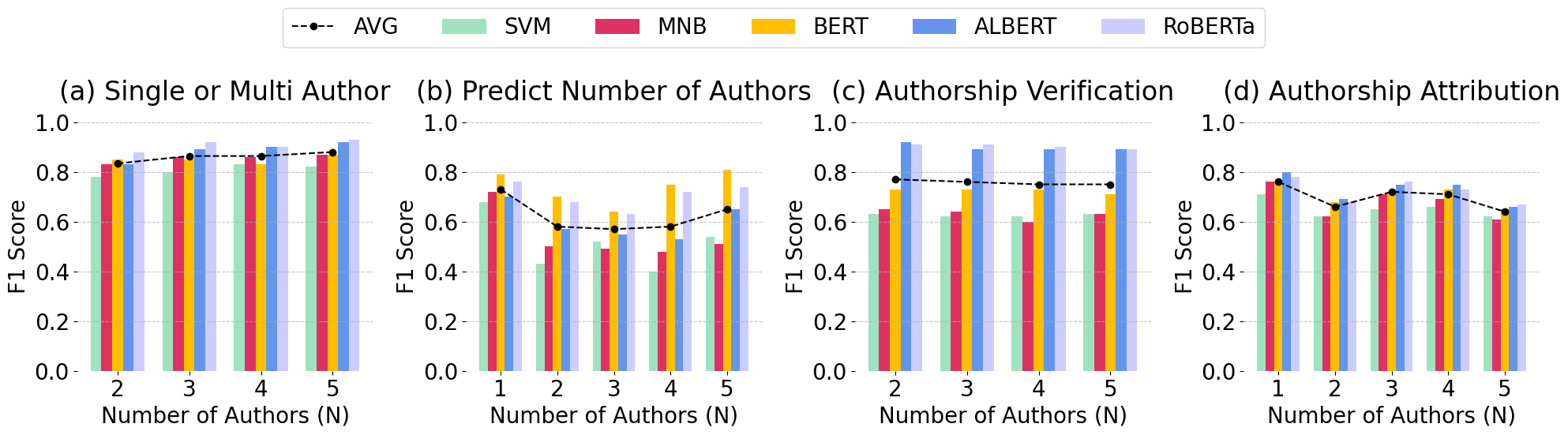}
  \caption{F1 scores for authorship-related tasks in the multi-LLM scenario, using different methods (color-coded) as the number of authors increases from N=1 to N=5.}
  \label{fig:authorship_results}
\end{figure*}

\section{Dataset Analysis}

We compare the LLM-generated single and multi-authored texts with the single-authored human written stories to study the relative quality of LLM generated stories as the number of authors present increases. To do this, we report the average and standard deviation of different measures such as the average number of (i) \textbf{words}, (ii) \textbf{sentences}, (iii) \textbf{vocabulary richness}, (iv) \textbf{percentage of stopwords}, (v) \textbf{readability scores}, (vi) \textbf{entropy}, and (vii) \textbf{coherence} scores using the TextDescriptives Library\footnote{\url{ https://github.com/HLasse/TextDescriptives}} for different numbers of authors ($N$) in Figure \ref{fig:data_stats}.  We measure (vii) \textbf{creativity} using OCSAI (Open Creativity Scoring with Artificial Intelligence) \cite{creativity_score} that provides text creativity scores using an LLM (GPT-4) fine-tuned on human annotations of creativity. Its high correlation with human judgments makes it a more reliable measure of overall creativity and divergent thinking than other methods that rely on semantic similarity. We used the Wilcoxon signed-rank test, a non-parametric method, to evaluate differences in text properties, including vocabulary richness, between human-written and LLM-coauthored stories (ranging from 1 to 5 LLM contributors). This test, with a significance level of p=0.05, accounts for non-normal distributions in the data.

From Figure \ref{fig:data_stats}, we see that across most measures, there isn't a significant deviation or decline as the number of authors is increased, i.e. collaboration or introducing multiple LLMs in the generation setting does not disturb the quality of the stories generated. In fact, the only statistically significant difference between LLM-coauthored and human-written stories in vocabulary richness measured by type-token-ratio (TTR) for $N \in [2,3,4]$. The lower scores in vocabulary richness, coherence, or creativity should not be considered inferior performance. Higher coherence scores in LLM-generated stories often originate from the tendency of LLMs to emphasize explicit logical flow and sentence-level similarity, producing semantically aligned and structured outputs. In contrast, human stories frequently employ nuanced storytelling techniques, including implicit connections, creative leaps, and varied narrative structures \cite{narration}, which, while enriching the narrative, may lead to lower scores in metrics like embedding-based coherence.

Human authors may exhibit lower vocabulary richness because, according to the theory of linguistic individuality \cite{indivduality}, individuals develop a mental ``key profile'' or preferred set of linguistic units, including specific words and phrases, that reflect their unique style and cognitive preferences. When writing stories, humans often consciously or unconsciously draw upon this limited set of vocabulary to maintain coherence, consistency, and fluency, which can restrict lexical variety compared to AI systems that leverage a broader and less constrained lexicon. For all other measures, our dataset's machine-generated stories follow similar distributions to human-written stories. Automated measures of readability and coherence utilized the human-generated stories as the reference text, and even for these measures, we see that collaboratively generated text scores do not deviate from the reference score distribution. This analysis indicates that LLMs are able to collaboratively generate stories without any significant changes in quality through sequential prompting. Detailed statistics for each LLM are provided in the Appendix (Table \ref{tab:per_author_stats}).

\begin{table}
    \centering
    \renewcommand{\arraystretch}{1.4}  
    \resizebox{\columnwidth}{!}{\begin{tabular}{>{\centering\arraybackslash}p{3cm}|>{\centering\arraybackslash}p{2cm}|>{\centering\arraybackslash}p{2cm}|>{\centering\arraybackslash}p{2cm}}
        \toprule
        \textbf{Negative Sample Source} & \textbf{Correct Wins} & \textbf{Neither Wins} & \textbf{Incorrect Wins} \\
        \midrule
        Different Story & \textbf{75\%} & 25\% & 0\% \\  \midrule
        Same Story & 35\% & \textbf{55.33\%} & 9.67\% \\
        \bottomrule
    \end{tabular}}
    \caption{Continutity evaluation of pairs of story parts using GPT-4o using different negative sample sources (either from within the same story or not). This table presents the percentage of story parts for which either only the correct part was evaluated as having continuity, or the negative sample (incorrect) or both. }
    \label{tab:continuity_results}
\end{table}

\subsection{Story Continuity}
An important aspect of collaborative storywriting is the notion of continuity. Particularly, we wanted to measure if the different story parts generated sequentially by different LLM authors followed a logical and cohesive plot. To evaluate this, we used a prompt-based evaluation using GPT-4o \cite{openai2024gpt4o} to discern if two consecutive story parts continue a story well or not. To do this, we develop two types of pairs of story parts: one pair containing the "Correct" or the positive sample which points to the actual next part in the story, or "Incorrect" which is the negative sample that is either a random story part drawn from a different story or from the same one (but not the true next one, any other part). In other words, for the negative samples taken from within the same story, we ensure to not use the correct consecutive part as a negative sample. An example of this setup is illustrated in Table \ref{tab:continuity_example}, where given a story part, there is an actual continuation ("Correct") and a random negative sample ("Incorrect") story part drawn from either the same story or a different one. The motivation for using the correct and incorrect pairs is to check if the evaluator (GPT-4o) gives high continuity scores to story parts (pairs) that were continuous and low continuity scores to story parts that were not consecutive. This type of A/B testing is a means of demonstrating that our LLM-based evaluation is reasonable/in accordance with expected results. We implicitly account for the biases as follows:

\begin{itemize}
    \item Length: we keep the length of story parts approximately similar using a word limit constraint in the generation prompts. This ensures that the lengths of the story parts in every pair being evaluated are similar.
    \item Position: The pairs being evaluated are drawn from all the possible positions in the text. In our evaluation, we present a general analysis of how continuous any two consecutive story parts are, irrespective of where they occur in the story, and present an aggregate over all such positions (beginning to end). 
    
    \item Self-Bias: We control for self-bias by ensuring that GPT-4o or a similar OpenAI model was not one of the LLMs being used for generation.     
\end{itemize}




\newcommand{\hlg}[2][green!40]{ {\sethlcolor{#1} \hl{#2}} }
\newcommand{\hlr}[2][red!40]{ {\sethlcolor{#1} \hl{#2}} }

\begin{table*}[t!]
\centering
\resizebox{\textwidth}{!}{\begin{tabular}{p{7cm}|p{7cm}|p{7cm}}
\toprule
Story Part &
  True Continuation &
  Random Negative Sample \\
\midrule
\hlg{Lisa} was born to a small village on the outskirts of a vast \hlg{forest}. All her life, she was a regular \hlg{village girl}, happily tending to the family’s crops and tending to any animal that came her way. The locals were fond of her, and she of them. On her eleventh birthday, she received a mysterious gift from the village elder. It was a small, silver orb, engraved with intricate runes. The orb seemed almost sentient, and as she looked upon its surface, a sense of calm washed over her like a gentle breeze. The days that followed were uneventful. Lisa went about her usual duties, always making sure her crops grew strong and healthy. She \hlg{rescued animals} in distress and cared for any sick or injured creatures she found. Life in the village was peaceful and serene. \hlg{Lisa never questioned why she was different}. She trusted her instincts and believed in the good in everyone. The villagers liked her for her pure heart, and her kindness was well-known throughout the land. &
  Yet, as the days turned to years, \hlg{Lisa couldn't shake off an inexplicable feeling of restlessness}. She spent long hours by the \hlg{forest}, watching the trees sway in the wind, and the sunlight filtering through their leaves. \hlr{On her eleventh birthday, she received a mysterious gift from the village elder - a beautifully crafted wooden box with intricate carvings of flowers and animals}. The box was locked, but the key was inside. Lisa opened it to find an old map of the forest, marked with symbols she had never seen before. Her heart raced as she stared at the map. She knew this was no ordinary gift. The village elder pulled her aside and revealed the truth - Lisa was the chosen one, destined to protect the forest from an ancient evil that had been sleeping for centuries. Lisa was taken aback, but she knew deep down that this was the reason for her restlessness. \hlg{She spent the next few weeks learning about the forest and the magic that resided within it}. &
  The \hlr{officer} stepped forward cautiously, his boots echoing through the \hlr{otherwise silent corridor}. He reached the end, where a single door stood slightly ajar, revealing a flickering light within. Drawing his weapon, he prepared for the unknown dangers that lay beyond. As he pushed the door open wider, he was greeted by a sight that chilled his blood. A room full of masked figures, their eyes glowing with a malevolent light, danced and swayed to an eerie tune. The officer's heart raced, and he knew that he had stumbled upon a \hlr{dangerous cult gathering}. He tried to retreat, but it was too late. The masked figures noticed him, and with a sinister cackle, they advanced towards him. The officer fired his weapon, hoping to scare them off, but his bullets seemed to have no effect. They closed in on him, their faces twisted into grotesque grins, and the officer knew that he was in grave danger. As the masked figures closed in on him, the officer felt a surge of determination. \hlr{He would not go down without a fight}. \\
 \bottomrule
  
\end{tabular}}
\caption{To evaluate the continuity of story parts, we sample two potential continuations to a story part, one correct and one incorrect. In this figure, we see an example of a story part on the left and two candidates to be evaluated for continuity. In \hlg{green} are some terms highlighted to show \hlg{continuity in topics, characters, and plot lines}, and in \hlr{red} are highlighted sections of the text that suggest \hlr{discontinuity}.}
 \label{tab:continuity_example}
\end{table*}

 We then use the following prompt structure to obtain continuity evaluations for both candidate next parts and then compare if the actual continuation was the one with a better continuation evaluation. We provide GPT-4o a pair of texts at a time (out of a total of 600 pairs) and ask it the following question: \textit{``Does Part 2 serve as a good continuation of Part 1 in terms of logical flow, coherence, and consistency? Please respond with Yes or No.''} GPT-4o generated the binary "yes/no" responses and an explanation. We then compare the percentage of story parts for which either the correct answer got a "Yes" and the incorrect one got a "No" response (Correct Wins), if both candidate continuations were evaluated the same (Neither Wins), or if the Incorrect answer got a "Yes" and the Correct one got a "No". We present the results in Table \ref{tab:continuity_results}. We see that when the negative story parts are sampled from different stories than the one that the first part belongs to, then there are 0 cases for which the incorrect part wins. This is an expected result, as story parts that belong to different stories would have different topics, plot lines, characters, and so on. Hence, detecting this discontinuity should not be hard, as is the case. An example of a case where Neither Wins and both candidate story parts get an evaluation as being discontinuous is provided in Table \ref{tab:continuity_example}. As illustrated by this example, it is crucial to note that our prompt encourages the model to evaluate the logical or common-sense flow of the stories. Hence, topic overlap alone was insufficient to mark two parts as continuous. In the harder setting, when the negative sample is also a story part from within the same story, we see that there is a marked increase in both the Neither Wins and Incorrect Wins scenarios (see Table \ref{tab:continuity_results}) where Incorrect Wins goes from 0 to 9.67\% and Neither Wins has an over 30\% increase. This is expected since many of the story parts were not the next immediate part but a few sequences apart. They all follow the same story and share some logical flow despite the distance in their occurrences in the story. For example, the fourth part of story might get a positive continuity evaluation, just as the second, third, and fifth would if compared with the first part of the story. Over 85\% of Neither Wins story pairs had a continuous evaluation for both candidates (142 out of 166). Thus, out-of-the-box LLMs are able to follow plot lines and logic even when continuing each others' partially written parts at any point in the story.

\section{Authorship Analysis: Extending PAN tasks for multi-LLM scenario}
Plagiarism Analysis, Authorship Identification, and Near-Duplicate Detection, known as PAN tasks \cite{pan_task}, have presented a persistent challenge, establishing benchmarks for analyzing multi-authored text among humans for more than 15 years. We extend the most common and repeated authorship-related tasks from the PAN multi-human-author task suite to the multi-LLM scenario. We then fine-tune and report performance using the following 5 baseline methods: Multinomial Naive Bayes (\textbf{MNB}) \cite{mnb}, Support Vector Machine (\textbf{SVM}) \cite{svm}, \textbf{BERT} \cite{bert}, \textbf{ALBERT} \cite{albert}, and \textbf{RoBERTa} \cite{roberta}. We used a 70:30 split of the dataset for fine-tuning and evaluation and aggregated performance over three random splits.

\subsection{Task 1: Is a story written by multiple authors or not?}
We randomly sample articles from the single-LLM authored stories i.e. $N=1$ as the negative class v/s articles from the multi-authored settings where $N \in [2,3,4,5]$ as the positive class. We sample from the single-LLM stories to keep the class distribution equal, based on the number of articles for each $N$. From Figure \ref{fig:authorship_results}(a), we see that for all methods, the performance at $N=5$ is higher than for $N=2$, gradually increasing with the value of $N$. Stories that have a higher number of authors are more distinct from single-authored ones. We conjecture that introducing more authors in the article might lead to more variations in the text, making stories with $N=5$ authors most easily distinguishable from stories without any such variations i.e. $N=1$.

\subsection{Task 2: How many authors have written a story?}
The second task is to predict the number of authors involved in generating a story. For the {\dataset} dataset, this means that class labels $\in [1,5]$. From Figure \ref{fig:authorship_results}(b) we see that the task of predicting exactly how many authors have co-written a story is easiest for $N=1$ in conjunction with findings from Task 1 that showed that multi-authored text can be more easily distinguished from single-authored text. Thus, here too it seems to be easiest to separate the single-authored texts from $N \geq 2$. However, for multi-authored stories, only BERT and RoBERTa perform better than other baselines (>0.72 F1), especially for $N \in [4,5]$. Overall, the performance across this task is low.

\subsection{Task 3: Authorship Verification}
This is a pair-wise sentence classification task where the goal is to predict if two consecutive sentences are written by the same author or not. For this task, we used all the sentences at LLM-LLM boundaries, that is the last sentence of part $i$ and the first sentence of part $i+1$. The negative class data samples were sampled as random pairs of consecutive sentences within each story part. From Figure \ref{fig:authorship_results}(c), we see that transformers-based fine-tuned methods perform well at this task (> 0.8 F1). 

\subsection{Task 4: Authorship Attribution}
Authorship Attribution involves predicting exactly who the author of a text article is. In the case of multi-LLM text, we design this task such that each data sample is written by a single author (does not contain any text windows that have an authorship change), and the classifier's task is to identify its author. From Figure \ref{fig:authorship_results}(d), we see that irrespective of the value of $N$, attribution performance is low (between 0.6 to 0.75 F1). We were expecting attribution to be easier the fewer the number of authors in the article since the length of the parts contributed by each author would be longer. But there does not seem to be any such correlation in our dataset i.e. length contributed by each author does not correlate with their detection. We provide more detailed results for each LLM or class-wise F1 scores in the Appendix (Table \ref{tab:aa_results}).

Thus, from the perspective of authorship tasks, for the multi-LLM scenario, Task 1 and Task 3 have higher performance (>=0.8 F1) and it is relatively easier to predict if a text is single or multi-authored and if two consecutive texts are written by the same author. , Task 2 and Task 4 are challenging even for fine-tuned LMs, which means predicting exactly \textbf{how many authors have contributed to a text} and exactly \textbf{who wrote each span of text} are extremely hard in this multi-LLM scenario, irrespective of how few or how many LLMs are collaborating.

\section{Conclusion}

We present {\dataset}, the first exclusively LLM-LLM or machine-machine collaborative story dataset, and demonstrate the tasks it enables. We find that LLMs are able to collectively generate creative stories at par with human-written stories via sequential prompting. Using this dataset, we demonstrate which multi-LLM authorship tasks are most challenging. Recent developments have significantly advanced LLM-assisted writing, sparking widespread discussions about the nature of authorship. Beyond using LLMs for paraphrasing, editing, and enhancing text, there exists an extreme scenario where text is generated entirely by multiple LLMs. Our work addresses this extreme case, raising several nuanced authorship concerns: Who should be considered the true creative source in such a situation? Should all LLMs involved be credited? Or should the human developers designing the prompts be acknowledged as the primary authors? Moreover, should the LLM that contributed the most—whether in terms of word count, narrative depth, or plot twists—be granted greater ownership? We will soon need ``Catch As Catch Can'' methods to not only find all points where authorship changes within an article but also simultaneously attribute each independent segment to the specific LLM author. As more and more LLMs are becoming easier to access, malicious actors could combine texts from different LLMs to evade automated and in-built misinformation flaggers, or students might circumvent credibility checks by having different LLMs write different sections of an academic article. We further discuss the real-world implications of such tasks in the Appendix (Section \ref {sec:real_world}). Thus, {\dataset} has been developed as a resource with long-form stories written by multiple LLMs to support the development and expansion of tasks and methods that can help address incoming challenges brought by LLM-LLM interactions.


\section*{Limitations}
Our work demonstrates one way of collecting a collaborative multi-LLM dataset. However, several variants are possible. Of course, as the LLM space is ever evolving, newer LLMs (e.g. Llama 3) became available as we were already collecting this dataset. Another aspect is that our dataset was collected in a uniform manner such that all LLMs contributed somewhat equal portions of text to a story. The next step would be to train a routing algorithm or a randomizer that could generate non-uniform collaborative texts. Our current analysis is unable to account for this setting and we leave this for future work. Additionally, the iterative generation process is resource-intensive and not easily scalable. We also acknowledge that LLM tasks beyond story writing are essential for a deeper understanding of how LLMs collaborate in open-ended generation tasks. 

\section*{Ethics Statement}
Using LLMs for creative story writing could relay some of the biases and harmful stereotypes present in the LLMs original training data since all our LLMs are trained on data from the internet. This is an important consideration before or during the dissemination of any such generated texts or stories. Transparency of the source of generated articles is important to avoid deception or wrongful content attribution. With creative writing tasks, it is also important to address any impact on creative professionals and guidelines to ensure that LLMs help enhance rather than undermine human creativity. We study LLM story-writing as a means to better prepare for a future of LLM-generated creative texts that might be misused in classroom settings, to manipulate public opinion on social media forums, and also to protect human writers against plagiarism amongst many other potential non-ethical usages. 

\section{Acknowledgments}
This work was in part supported by NSF awards \#1934782 and \#2114824. Some of the research results were obtained using computational resources provided by CloudBank (\url{https://www.cloudbank.org/}), which was supported by NSF award \#1925001.

\bibliography{main}
\appendix
\clearpage
\section{Appendix}


\subsection{LLM Prompting} \label{sec:llm_prompting}
For all $N \geq 2$, we provided the summary of the story so far as an input in the prompt. To make sure that our story parts had smooth continuity, we also used the last sentences of the story so far as input. This ensured that the generating LLM has access to the last sentence and the overall storyline to continue the story as seamlessly as possible. This second input is denoted as ``last sentence from previous part'' in Table \ref{tab:prompts}. Other than this, our prompt for the three types of story sections only differed in the instruction of writing either the ``beginning'', ``continue'', or ``conclude'' the storyline so far. We also had to add an instruction to stop the LLMs from generating any additional instructions as from our pilot study, we found that some LLMs (Orca and Llama) would often first generate a rephrasing or more detailed version of our instruction before generating the actual story content. 

\subsection{Dataset Cleaning} \label{sec:data_cleaning}
For each prompt, we gave each LLM 20 maximum attempts to re-generate that particular story part if it fell 15 or more words shorter than the goal length in the previous iteration. Despite this, we had instances of very short story parts that would have made the average article length too short or led to a very skewed representation of one LLM v/s the rest. Thus, we discarded such stories. Additionally, we were able to notice two formatting peculiarities for Gemma and LLama. Particularly, Gemma's story parts often began with a short title for the section it was to generate surrounded by ``\#\#\#'' for example ``\#\#\# The return of the Jedi \#\#\#''. Llama on the other hand was appending a ``The end'' whenever it was its turn to write the ending part of a story. We removed all cases of these two substrings using regular expressions search and deletion as a means to unify the flow of the story across all LLMs and to make sure particular LLMs weren't identifiable only due to such formatting details. We also removed all extra spaces from the stories and any repetitions of the instructions in rare cases.

\begin{table*}[ht!] 
\centering
    \resizebox{\textwidth}{!}{\begin{tabular}{m{4cm}|m{8cm}|m{8cm}}
    \toprule
        \multicolumn{1}{l}{PAN Task equivalent} &  \multicolumn{1}{c}{Task Description: Multi-LLM scneario} &  \multicolumn{1}{c}{Real-world Implications} \\ \midrule
        Predict multi-author or not & To determine if a text includes content from multiple LLMs or not & Credit Assignment and Intellectual Property (IP) regulation \\ \midrule
        Predict number of authors & To predict the number of LLMs involved in writing an article & Keeping track of LLM-LLM agent interactions in growing open-source market \\ \midrule
        Author Verification & To detect when authorship switches between LLMs & To detect perjury, misinformation injection, falsifying editing in news articles, and text obfuscation \\ \midrule
        Authorship Attribution & Predicting who wrote each text segment? & Plagiarism detection \\ \midrule
        Style Change Detection and Attribution & Finding all positions in the text where authorship changes and who wrote each segment & Classroom settings: Academic Integrity, detecting use of multiple open-source and free-to-use LLMs to surpass detection methods \\ 
        \bottomrule
    \end{tabular}}
\caption{Real-world implications of the tasks involved in understanding LLM-LLM collaboration for writing tasks}
\label{tab:implications}
\end{table*}

\begin{table*}[t!]
  \centering
    \resizebox{\textwidth}{!}{\begin{tabular}{|ll|ccccc|}
        \toprule
        Feature & Author & K=1 & K=2 & K=3 & K=4 & K=5 \\ 
        \midrule
        \# Words & Gemma & 172.97 ± 16.47 & 157.17 ± 36.22 & 124.51 ± 46.13 & 129.76 ± 47.16 & 133.75 ± 46.90 \\ 
        & Llama & 172.51 ± 19.91 & 170.88 ± 13.28 & 173.35 ± 15.29 & 174.76 ± 16.18 & 172.23 ± 19.33  \\ 
        & Mistral & 177.25 ± 12.61 & 182.24 ± 12.88 & 178.09 ± 22.71 & 178.82 ± 19.47 & 178.15 ± 22.69 \\ 
        & Olmo & 168.01 ± 8.69 & 197.91 ± 18.93 & 194.89 ± 23.64 & 192.64 ± 26.60 & 191.41 ± 30.28 \\ 
        & Orca & 174.45 ± 22.20 & 175.61 ± 10.42 & 178.11 ± 15.62 & 178.01 ± 14.37 & 177.89 ± 16.57 \\ 
        \midrule
        Lexical Diversity & Gemma & 0.67 ± 0.04 & 0.67 ± 0.05 & 0.70 ± 0.07 & 0.69 ± 0.07 & 0.68 ± 0.06 \\ 
         & Llama & 0.60 ± 0.05 & 0.61 ± 0.05 & 0.59 ± 0.05 & 0.58 ± 0.05 & 0.58 ± 0.06 \\ 
         & mistral & 0.64 ± 0.04 & 0.62 ± 0.04 & 0.61 ± 0.05 & 0.61 ± 0.04 & 0.61 ± 0.05 \\ 
         & Olmo & 0.62 ± 0.05 & 0.57 ± 0.06 & 0.58 ± 0.06 & 0.58 ± 0.06 & 0.57 ± 0.07 \\ 
         & Orca & 0.62 ± 0.05 & 0.62 ± 0.04 & 0.61 ± 0.04 & 0.61 ± 0.05 & 0.60 ± 0.05 \\ 
        \midrule 
        Readability & Gemma & 75.95 ± 8.53 & 77.13 ± 9.70 & 75.13 ± 12.29 & 74.11 ± 11.81 & 73.28 ± 12.96 \\ 
         & Llama & 83.11 ± 8.35 & 82.61 ± 8.50 & 82.75 ± 9.00 & 80.88 ± 10.10 & 80.13 ± 9.19 \\ 
         & Mistral & 81.04 ± 8.58 & 83.99 ± 8.40 & 81.59 ± 10.07 & 82.02 ± 9.09 & 79.91 ± 9.44 \\ 
         & Olmo & 80.78 ± 9.01 & 83.31 ± 9.87 & 81.41 ± 9.81 & 80.55 ± 10.73 & 80.45 ± 10.20 \\ 
        & Orca & 83.08 ± 8.54 & 82.51 ± 8.45 & 80.86 ± 9.65 & 79.95 ± 9.97 & 79.51 ± 9.80 \\
         \midrule
        Coherence & Gemma & 0.49 ± 0.07 & 0.47 ± 0.08 & 0.47 ± 0.08 & 0.47 ± 0.08 & 0.48 ± 0.08 \\ 
         & Llama & 0.44 ± 0.08 & 0.47 ± 0.08 & 0.47 ± 0.08 & 0.47 ± 0.08 & 0.49 ± 0.08 \\ 
         & mistral & 0.46 ± 0.07 & 0.44 ± 0.07 & 0.45 ± 0.07 & 0.45 ± 0.07 & 0.46 ± 0.07 \\ 
         & Olmo & 0.47 ± 0.08 & 0.43 ± 0.07 & 0.45 ± 0.08 & 0.45 ± 0.08 & 0.45 ± 0.08 \\ 
         & Orca & 0.44 ± 0.07 & 0.45 ± 0.08 & 0.45 ± 0.08 & 0.46 ± 0.08 & 0.47 ± 0.07 \\ 
         \bottomrule
    \end{tabular}}
\caption{Descriptive Statistics or Features for stories generated by different authors for different parts of the stories. Here, ``K'' represents the part of the story written, i.e. K=1 corresponds to the first part of the story, K=2 referees to the second part, and so on.}
\label{tab:per_author_stats}
\end{table*}

\begin{table}[t!]
\centering
\resizebox{0.5\textwidth}{!}{\begin{tabular}{|l|ccccc|c|}
\multicolumn{7}{c}{Task 4: Authorship Attribution} \\
\toprule
\multicolumn{7}{c}{N=1} \\
\midrule
Method & Orca & Olmo & Llama & Mistral & Gemma & AVG \\
\midrule
MNB & - & 0.70 & \underline{0.71} & 0.64 & \textbf{0.99} & 0.76 \\
SVM & - & 0.61 & 0.68 & 0.58 & \underline{0.97} & 0.71 \\
BERT & - & 0.70 & \underline{0.71} & 0.64 & \textbf{0.99} & 0.76 \\
ALBERT & - & \textbf{0.78} & \textbf{0.73} & \textbf{0.70} & \textbf{0.99} & \textbf{0.80} \\
RoBERTa & - & \underline{0.73} & 0.70 & \underline{0.68} & \textbf{0.99} & \underline{0.78} \\
\midrule
\multicolumn{7}{c}{N=2} \\
\midrule
Method & Orca & Olmo & Llama & Mistral & Gemma & AVG \\
\midrule
MNB  & 0.49 & 0.51 & 0.52 & 0.51 & 0.92 & 0.62 \\
SVM & 0.51 & 0.55 & 0.54 & \textbf{0.59} & 0.79 & 0.62 \\
BERT & \underline{0.54} & 0.54 & \textbf{0.63} & \underline{0.58} & \underline{0.95} & \underline{0.68} \\
ALBERT & \textbf{0.56} & \underline{0.58} & \textbf{0.63} &\textbf{0.59} & \textbf{0.96} & \textbf{0.69} \\
RoBERTa & 0.49 & \textbf{0.62} & \underline{0.60} & 0.56 & 0.94 & \underline{0.68} \\
\midrule
\multicolumn{7}{c}{N=3} \\
\midrule
Method & Orca & Olmo & Lama  & Mistral & Gemma & AVG  \\
\midrule
MNB    & - & 0.60 & 0.67 & 0.63 & 0.94 & 0.71 \\
SVM    & - & 0.57 & 0.65 & 0.57 & 0.82 & 0.65 \\
BERT   & - & 0.58 & \underline{0.69} & \underline{0.67} & \underline{0.95} & 0.72 \\
ALBERT & - & \underline{0.64} & \textbf{0.71} & \textbf{0.68} & \underline{0.95} & \underline{0.75} \\
RoBERTa& - & \textbf{0.71} & \textbf{0.71} & \underline{0.67} & \textbf{0.96} & \textbf{0.76} \\
\midrule
\multicolumn{7}{c}{N=4} \\
\midrule                       
Method & Orca & Olmo & Llama & Mistral & Gemma & AVG  \\
\midrule
MNB    & - & 0.58 & 0.65 & 0.63 & \underline{0.91} & 0.69 \\
SVM    & - & 0.58 & 0.67 & 0.59 & 0.80 & 0.66 \\
BERT   & - & \underline{0.59} & \underline{0.70} & \underline{0.68} & \textbf{0.93} & \underline{0.73} \\
ALBERT & - & \textbf{0.66} & \textbf{0.73} & \textbf{0.70} & \textbf{0.93} & \textbf{0.75} \\
RoBERTa& - & \textbf{0.66} & 0.68 & 0.64 & \textbf{0.93} & \underline{0.73} \\
\midrule
\multicolumn{7}{c}{N=5} \\
\midrule                             
Method & Orca & Olmo & Llama & Mistral & Gemma & AVG  \\
\midrule
MNB    & 0.54 & 0.54 & 0.56 & 0.53 & 0.86 & 0.61 \\
SVM    & 0.56 & \underline{0.61} & 0.60 & 0.54 & 0.79 & 0.62 \\
BERT   & \textbf{0.60} & 0.57 & \underline{0.62} & 0.54 & \textbf{0.93} & 0.65 \\
ALBERT & \underline{0.58} & 0.55 & \textbf{0.65} & \textbf{0.62} & \underline{0.92} & \underline{0.66} \\
RoBERTa& 0.56 & \textbf{0.69} & 0.61 & \underline{0.58} & \underline{0.92} & \textbf{0.67} \\
\bottomrule
\end{tabular}}
\caption{F1-scores for identifying the author of story parts across articles written by different numbers of authors. The 5 columns show each of the labels or authors. AVG denotes average F1-scores across all authors. Best performing method is in \textbf{bold} and second highest \underline{underlined}.}
\label{tab:aa_results}
\end{table}

\section{Authorship Analysis: Real world implications} \label{sec:real_world}

This work has profound implications for various stakeholders in the burgeoning socio-technical system of generative AI \cite{ip1, ip2, ip3, legal_impact}. Our research introduces authorship-related tasks using {\dataset}, which can help address these concerns by accurately discerning the usage of multiple LLMs in texts (summarized in Table \ref{tab:implications}. Our extension of PAN-inspired authorship tasks is closely linked to real-world implications, as follows:

\paragraph{Task 1: Predict multi-author or not} In the rapidly expanding and fiercely competitive market for LLMs, the ownership of content and the ability to prove the origins of creative work are becoming increasingly crucial. As the market evolves, closed-source LLMs are implementing stricter regulations and demanding credit assignment under various distribution licensing norms. In this context, the capacity to demonstrate that a text incorporates generated output from multiple LLMs is essential. This capability can effectively prevent any single stakeholder or developer from erroneously claiming exclusive rights to the content, thereby bolstering the defense against wrongful intellectual property (IP) claims.

\paragraph{Task 2: Predict number of authors} Predicting the exact number of LLMs involved in the writing process can help keep track of the frequency and extent to which LLMs are used collaboratively, as more and more models enter the open-source market. This is essential to understand whether such usage improves task performance or introduces inefficiencies beyond a certain threshold.  Understanding the optimal number of LLMs or the degree to which LLMs can leverage each other's strengths in writing tasks is vital. It ensures effective collaboration without unnecessary complexity, maximizing the benefits of combined model capabilities while avoiding overkill and collaboration for its own sake.

\paragraph{Task 3: Author Verification} With LLMs increasingly paraphrasing and editing each other's texts, it becomes crucial to identify which spans were generated by different LLMs. Consider a scenario where a news article is paraphrased by one LLM and subsequently edited by another, with the latter introducing fallacies or misinformation. In such cases, discerning the contributions of each LLM is essential for identifying malicious LLM agents or the infiltration of critical content, such as media and news articles. This capability has significant applications, including detecting perjury and combating the adversarial obfuscation of text, thus maintaining the integrity and reliability of information. 

\paragraph{Task 4: Authorship Attribution} Identifying the exact LLMs responsible for authoring a text is crucial for detecting and addressing plagiarism. This is particularly important in academic settings, where students might use closed-source LLMs without complying with content ownership and usage declaration regulations. This is possible also in cases where content from one LLM is being posed as that from another to claim higher ability or quality. An example of such a situation might be in a bid to motivate financial investors hoping to monetize and utilize LLMs for specific domains (such as medical applications, educational tools, and creative content generation).

\end{document}